\documentclass[final]{cvpr}

\usepackage{times}
\usepackage{epsfig}
\usepackage{graphicx}
\usepackage{amsmath}
\usepackage{amssymb}
\usepackage{bbm}
\usepackage{color}
\usepackage{multirow}
\usepackage[table,xcdraw]{xcolor}

\usepackage[pagebackref=true,breaklinks=true,colorlinks,bookmarks=false]{hyperref}

\def\cvprPaperID{6299} 
\def\confYear{CVPR 2021}

\begin{document}

\title{ 
Modeling Multi-Label Action Dependencies for Temporal Action Localization
}

\author{Praveen Tirupattur\\
University of Central Florida\\
{\tt\small praveentirupattur@knights.ucf.edu}
\and
Kevin Duarte\\
University of Central Florida\\
{\tt\small kevin\_duarte@knights.ucf.edu}
\and

Yogesh Rawat\\
University of Central Florida\\
{\tt\small yogesh@crcv.ucf.edu}

\and
Mubarak Shah\\
University of Central Florida\\
{\tt\small shah@crcv.ucf.edu}
}

\maketitle

\begin{abstract}
   Real-world videos contain many complex actions with inherent relationships between action classes. In this work, we propose an attention-based architecture that models these action relationships for the task of temporal action localization in untrimmed videos.
   As opposed to previous works that leverage video-level co-occurrence of actions, we distinguish the relationships between actions that occur at the same time-step and actions that occur at different time-steps (i.e. those which precede or follow each other).
   We define these distinct relationships as action dependencies. 
   We propose to improve action localization performance by modeling these action dependencies in a novel attention-based Multi-Label Action Dependency (MLAD) layer. 
   The MLAD layer consists of two branches: a Co-occurrence Dependency Branch and a Temporal Dependency Branch to model co-occurrence action dependencies and temporal action dependencies, respectively. 
   We observe that existing metrics used for multi-label classification do not explicitly measure how well action dependencies are modeled, therefore, we propose novel metrics that consider both co-occurrence and temporal dependencies between action classes. Through empirical evaluation and extensive analysis, we show improved performance over state-of-the-art methods on multi-label action localization benchmarks (MultiTHUMOS and Charades) in terms of f-mAP and our proposed metric.
   
\end{abstract}


\section{Introduction}
\label{sec:intro}

\begin{figure}[t]
\begin{center}
\includegraphics[width=\linewidth]{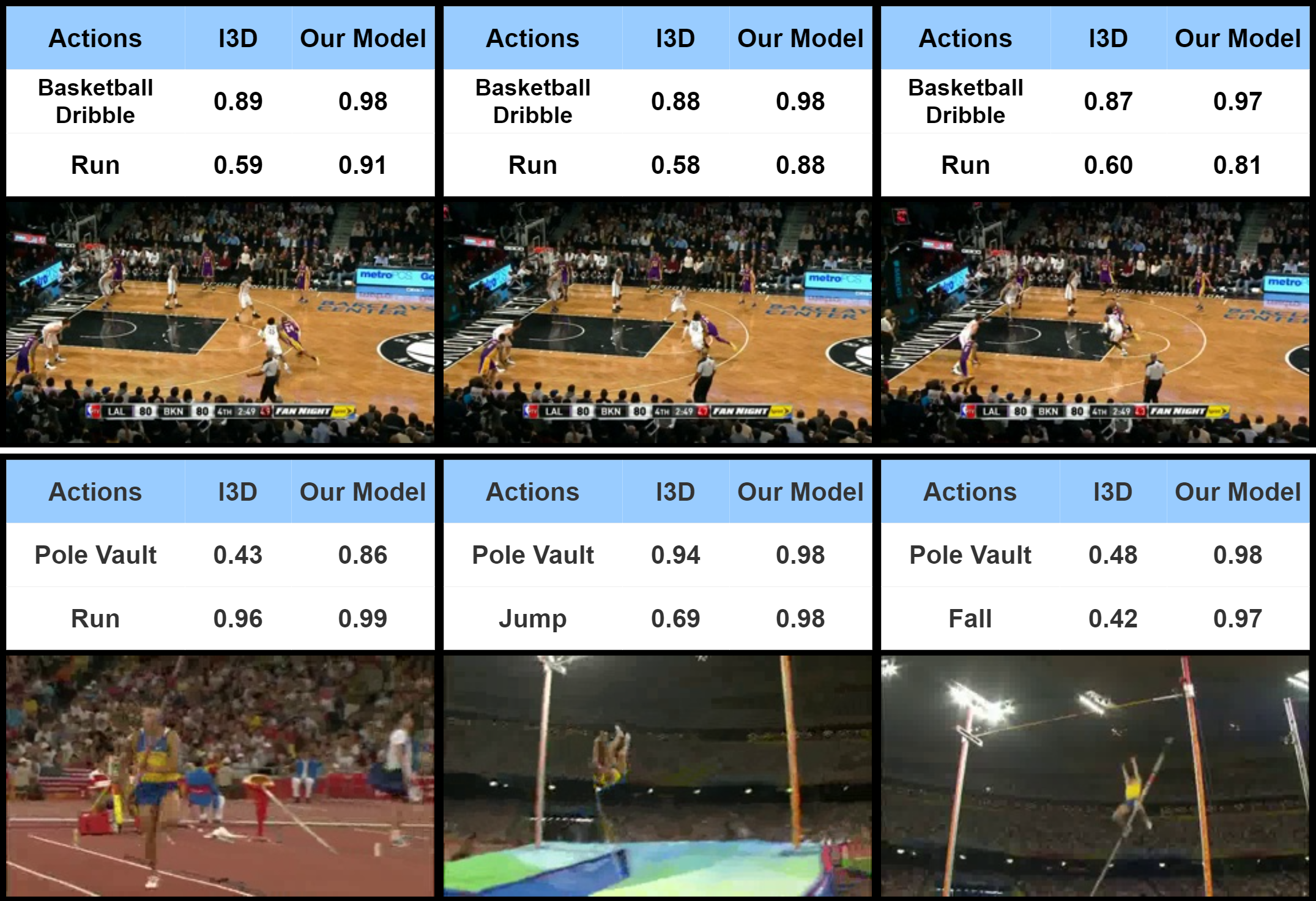}
\end{center}
\caption{Two action sequences from the MultiTHUMOS dataset. The first sequence (top) shows action dependencies within a given time-step: ``Basketball Dribble" and ``Run" co-occur. The bottom sequence shows action dependencies across time-steps: ``Fall" follows ``Jump". The table above each frame shows the comparison of probability scores predicted by our model with the I3D baseline for each action class present at that time-step. Modeling both types of dependencies is beneficial for correctly detecting actions. 
}
\label{fig:motivation}
\end{figure}

Understanding and localizing actions in complex video sequences is a heavily researched problem in computer vision. The task of action localization in the untrimmed video involves predicting the action, or actions, present at each time-step of the video sequence. Several works propose top-down methods, that propose temporal regions of a video which are then classified and refined \cite{caba2016fast, escorcia2016daps, shou2016temporal, buch2017sst, gao2017turn, zeng2019graph}. Other approaches produce bottom-up predictions for each time-step directly from the frame-level or clip-level features \cite{lin2018bsn, lin2019bmn, liu2019multi, shou2017cdc, long2019gaussian}. Recent bottom-up methods tend to perform best on the multi-label case, where multiple actions can be present within the same time-step.

Although these works achieve strong multi-label action localization performance, they do not explicitly model the relationships between the different action labels, which can be extremely useful for determining the presence or absence of classes within a video. Previous works have used label co-occurrence to improve performance on image classification \cite{wen2020multilabel, yazici2020orderless, chen2020label}, and video action recognition \cite{bhattacharya2016covariance, modiri2014video}. However, the later works measure the video-level co-occurrence of actions, which does not differentiate between actions that occur {\em within the same time-step} and {\em across different time-steps}. This may be acceptable when the problem is video-level single-label action recognition, but when the task is to temporally localize multiple actions  (as is the case with multi-label temporal action localization) the distinction between these co-occurrences allows for more fine-grained modeling of action relationships. We define these distinct action class relationships as \textit{action dependencies}.

Videos contain two types of action dependencies: i) \textit{co-occurrence dependencies}, involving actions that occur at the same time (this is most analogous to object class co-occurrence within images), and ii) \textit{temporal dependencies}, involving actions that precede or follow each other. To illustrate, consider Figure \ref{fig:motivation} showing sample frames from pole vault and basketball videos. An example of a co-occurrence dependency is present in the first video snippet: the action ``run" often occurs with the action ``basketball dribble" in a basketball game, so the presence of one action gives additional prior information about the other. The second video snippet is an example of a temporal dependency. Using the available label information from the previous clips, one could infer the label following ``jump" to be ``fall" in the final clip even without visual or motion features corresponding to the person performing the action.

In this work, we present a method that leverages both action dependency types to improve learned feature representations for the task of multi-label temporal action detection. We propose an attention-based layer to refine class-level features based on these dependencies. Co-occurrence dependencies are modeled by refining action features based on the presence, or absence, of other actions within a time-step; temporal dependencies are modeled by refining features based on all the time-steps of an input video sequence. In both cases, attention maps are generated which allows for improved interpretability of our model. Differing from action recognition methods that employ class co-occurrence \cite{modiri2014video}, our approach does not require a ground-truth action co-occurrence matrix, but rather {\em learns} action dependencies from the training data.

To better understand how our approach models action dependencies, we present novel metrics for evaluating temporal action localization methods. Whereas previous multi-label evaluation methods, like mean average precision (mAP) and F1-score, tend to evaluate per-frame class performance independently, our proposed action-conditional precision and recall metrics explicitly measure how well pair-wise class/action dependencies are modeled both within a time-step and through different time-steps. Our proposed metrics are general - they can be applied to both images and videos by measuring performance on both co-occurrence and temporal action dependencies.


Our main contributions include the following:
\begin{itemize}
    \item We present a novel network architecture that models both co-occurrence action dependencies and temporal action dependencies.
    \item We propose multi-label performance metrics to measure a method's ability to model class co-occurrence across time-steps as well as within a time-step.
    \item We evaluate the proposed approach on two large scale publicly available multi-label action datasets, outperforming existing state-of-the-art methods.
\end{itemize}


\section{Related Work}
In recent years, temporal action localization research has received a lot of interest. In general, approaches for temporal action localization are broadly classified into top-down, bottom-up, and end-to-end. Top-down approaches \cite{caba2016fast, escorcia2016daps, zeng2019graph, buch2017sst}, start with candidate proposals and refine them to achieve the final temporal boundaries. These approaches perform well, but are often slow and suffer from over-generated proposals and rigid boundaries. Bottom-up approaches \cite{lin2018bsn, lin2019bmn, liu2019multi} start with frame-level or clip-level predictions for each action class and combine the individual scores to generate the final temporal boundaries. End-to-end approaches \cite{yeung2016end,lin2017single,buch2019end} integrate proposal generation and classification steps. These approaches are proposed to solve temporal action localization with non-overlapping instances and do not consider the relationships between action classes. In a multi-label setup, there are overlapping instances of different actions leading to parts of the video corresponding to multiple classes.

Multi-label classification has been studied in both images \cite{wang2016cnn, guo2019visual, chen2019multi, durand2019learning} and videos \cite{xie2017deep, kim2018temporal, na2017encoding} 
In the image domain, it has been shown that leveraging relationships between classes help improve classifier performance. 
Some works \cite{chen2019multi, li2014multi, li2016conditional, chen2020label, you2020cross} use probabilistic graphical models to incorporate label relationships by formulating this task as a structural inference problem. Others \cite{wang2016cnn, liu2017semantic, jin2016annotation, wang2017multi, yazici2020orderless} use spatial attention  with recurrent neural networks to model the label co-occurrence. 
In \cite{yeh2017learning, wen2020multilabel, bhatia2015sparse} image features and label domain data are projected to a common latent space to learn the label correlations. 

Videos introduce additional temporal relationships between labels which are crucial for multi-label temporal action localization. Most previous 
approaches \cite{gan2015devnet, campos2017skip, long2018multimodal, long2018attention, piergiovanni2019temporal} for multi-label temporal action localization neither consider the label co-occurrence nor the temporal relationships between the labels. Recently, some works have explicitly modeled temporal relationships between action labels \cite{santa2020inferring, piergiovanni2020differentiable}. 
The idea of learning a differentiable grammar to model high-level temporal structure and relationships between multiple action classes was introduced in \cite{piergiovanni2020differentiable} for the first time. 
In \cite{santa2020inferring}, a framework is presented to learn temporal ordering between atomic actions to detect complex activities in videos, by using regular expressions to express the temporal composition of atomic actions.
To the best of our knowledge, no existing works explicitly model {\em both} the co-occurrence and temporal dependencies between action classes. We propose a bottom-up approach that models the co-occurrence and temporal dependencies using attention to improve multi-label action localization performance.  
 

\begin{figure*}[t!]
\begin{center}
\includegraphics[width=\linewidth]{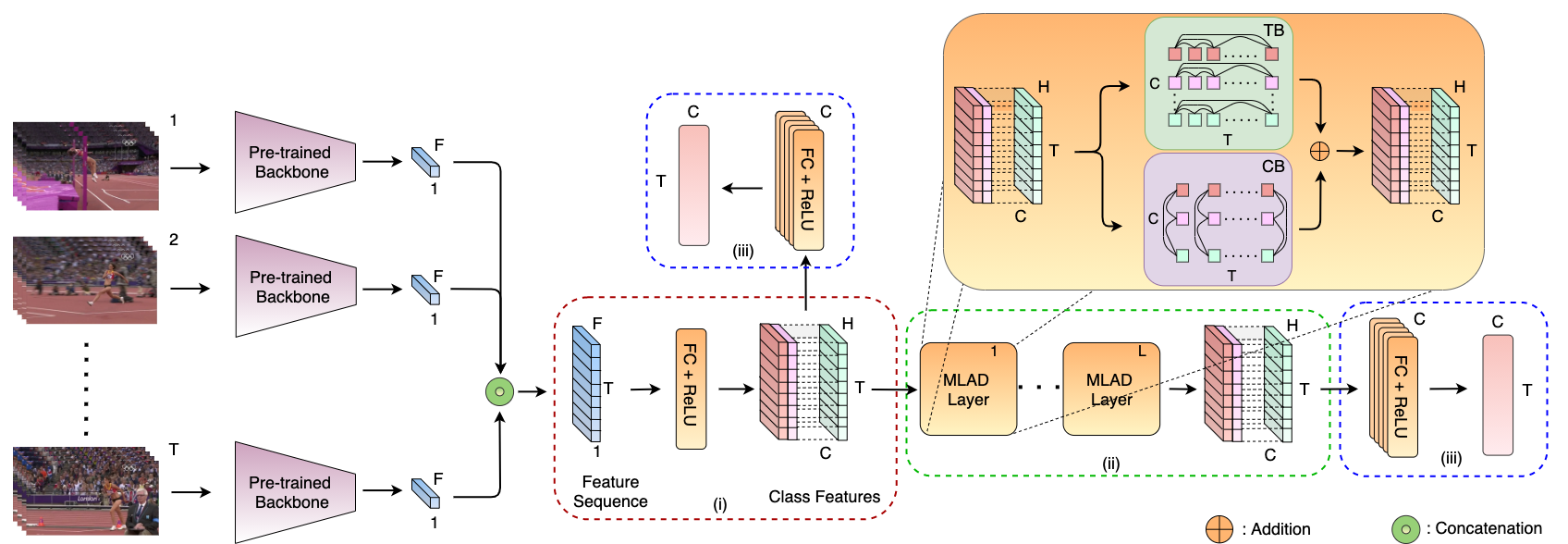}
\end{center}
\caption{Architecture of our proposed approach. 
Input to our model is a sequence of features ($T \times F$), extracted using a pre-trained backbone. Our proposed architecture process these features in three steps. First,  it learns class-specific features ($T \times H$) for each class  $C$ (shown in block (i)). Second, it  refines these class-specific features using one or more of the attention-based Multi-Label Action Dependency (MLAD) layers  (shown in block (ii)). Third, it classifies the features using individual classification layers for each class, and output class probabilities for each time step ($T \times C$)  (shown in block (iii)). 
} 
\label{fig:approach}
\end{figure*}

\section{Approach}
In this section, we first present the formulation of the multi-label temporal action localization problem. Then, we describe our proposed network. It consists of three main parts: (i) class-level feature extraction, (ii) feature refinement by using our MLAD layer 
which models both types of action dependencies, and (iii) a classification step to transform the refined features into class probabilities. The network architecture is depicted in Figure \ref{fig:approach}.

\paragraph{Problem Formulation} 
The problem of multi-label temporal action localization involves classifying all activities occurring throughout a video at each time-step. Formally, in a feature sequence of length $T$, each time-step $t=1,...,T$ contains a ground-truth action label $y_{t,c} \in \{0,1\}$, where $c=1,...,C$ is the action class. Given a feature vector of length $F$, $x_t \in \mathbb{R}^{F}$, for each time-step, an activity detection network predicts class probabilities $\tilde{y}_{t,c} \in \left[0,1\right]$.

\paragraph{Class-level Feature Extraction}
The input to our network is a series of feature vectors $x_t$. Since these features contain global representations (either frame-level or video-clip level, when obtained from 2D-CNN encoders and 3D-CNN encoders, respectively), we convert them to class-level representations. This nonlinear transformation is as follows:
\begin{equation}
\label{eq:class-feats}
    f_{t,c} = \text{ReLU} \left( W_c^T x_t + b_c \right),
\end{equation}
where $W_c$ and $b_c$ are learned weights for each class $c$. These $H$-dimensional vectors contain information pertinent to a given action at each time-step.

\subsection{MLAD Layer}
\begin{figure}[t!]
\begin{center}
\includegraphics[width=\linewidth]{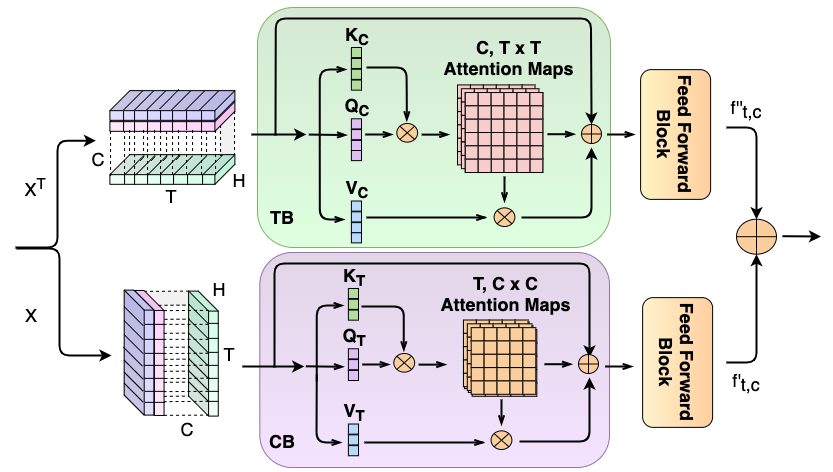}
\end{center}
\caption{\textbf{MLAD Layer}. Given class-specific features ($C \times H$) for each time-step $T$, this layer refines the features by modeling action dependencies with attention. The upper Temporal Dependency Branch (TB) models dependencies across time-steps (temporal dependencies) for each class and the lower Co-occurrence Dependency Branch (CB) models dependencies between classes within each time-step (co-occurrence dependencies). 
} 

\label{fig:mad}
\end{figure}

We propose a layer that can use these class-level features and model the relationships between the various action classes across time. One approach would be to use a fully-connected graph-based \cite{chen2019multi} or attention-based \cite{yan2019multi} network to learn the relationships between the feature vectors. This, however, would lead to $CT\times CT$ connections, which would be extremely inefficient when either the number of classes, $C$, or the number of time-steps, $T$, becomes large. Instead, we propose an efficient attention-based Multi-label Action Dependency (MLAD) layer which decomposes this operation into $C\times C$ and $T\times T$ sets of connections.
The MLAD layer contains two branches - the Co-occurrence Dependency Branch (CB) and Temporal Dependency Branch (TB) - which model their corresponding action dependencies and refine the input class-level features. Refer to Figure \ref{fig:mad} for the architecture of the MLAD layer.

\paragraph{Co-occurrence Dependency Branch (CB)} The CB models the relationships between actions within a given time-step. For each time-step, a self-attention operation \cite{vaswani2017attention} is performed across all classes. At each time step, $t$, input features generate a set of query, key, and value tensors ($Q_t,K_t, V_t$), each with dimension $\mathbb{R}^{C \times H}$. Then, a $C \times C$ attention matrix,  $A^{(t)}$, is  obtained as follows:
\begin{equation}
    A^{(t)} = \text{softmax} \left( \frac{Q_tK_t^T }{\sqrt{H}} \right).
\end{equation}
This attention matrix contains the relevance of each class for the classification of another class. For example, $A^{(t)}_{ij}$ denotes the relevance of class $j$ in the classification of class $i$ at time-step $t$; if these two classes co-occur within the same time-step often, then $A^{(t)}_{ij}$ should be large, otherwise, it will have a value close to 0. With this attention matrix, we obtain a refined set of class-level features that take into account the presence (or absence) of other classes within the time-step as follows:
\begin{equation}
f_{t, c}' = A^{\left(t\right)}V_t.
\end{equation}

\paragraph{Temporal Dependency Branch (TB)} The TB models actions' temporal dependencies. For each class, $c$, a new set of query, key, and value tensors ($Q_c,K_c, V_c$) are created with dimension $\mathbb{R}^{T \times H}$. The self-attention operation is performed across time:
\begin{equation}
    A^{(c)} = \text{softmax} \left( \frac{Q_cK_c^T }{\sqrt{H}} \right).
\end{equation}
Here, $A^{(c)}$ is a $T \times T$ attention matrix, where $A^{(c)}_{kn}$ denotes the importance of time-step $n$ in the classification of the given class, $c$, at time-step $k$. The refined features are obtained as follows:
\begin{equation}
f_{t, c}'' = A^{\left(c\right)}V_{c}.
\end{equation}
This branch incorporates information from all time-steps, producing more temporally coherent features and predictions. When the TB is used in conjunction with the CB, the MLAD layer can model both types of action dependencies.

\paragraph{Merging Branches and Classification} We merge the different sets of refined features ($f_{t, c}'$ and $f_{t, c}''$) to obtain a combined output representation. The trivial approaches for merging would be element-wise summation or concatenation followed by an MLP to reduce dimensionality. We propose to learn the amount of information which is used from each module; the module learns a value, $\alpha\in \left[0, 1\right]$, that is used to merge the outputs to compute combined features,  $g_{t, c}$, as follows:
\begin{equation}
    g_{t, c} = \alpha f_{t, c}' + \left( 1-\alpha \right) f_{t, c}''.
\end{equation}
We find that the use of the learned $\alpha$ term leads to improvement in performance when compared to element-wise averaging. The improved class-level feature representation, $g_{t, c}$, is either passed as an input to additional MLAD layers or used to produce a final classification output. This is performed by the transformation 
\begin{equation}
\label{eq:classifier-layer}
    \hat{y}_{t,c} = \sigma \left( W^T_c g_{t, c} + b_c \right),
\end{equation}
where $W_c \in \mathbb{R}^{d_k \times 1}$ and $b_c \in \mathbb{R}$ are learned weights and $\sigma$ is the logistic sigmoid function.

\section{Action Dependency Metrics}
The problem of multi-label temporal action localization consists of predicting the action, or actions, occurring at each time-step of a video.
The standard metric for evaluating temporal action localization, f-mAP, treats each time-step as an individual sample, measures the performance of each class independently, and averages their scores; it does not explicitly measure if models learn the relationships between these classes. This issue is not unique to f-mAP. Other multi-label classification metrics \cite{lei2015multi, tsoumakas2010random, schapire2000boostexter, wu2017unified} do not consider the relationships between different classes or time-steps, which makes them unsuitable to evaluate how well action dependencies are modeled. To this end, we propose new action localization metrics that measure a method's ability to model both co-occurrence dependencies and temporal dependencies.


For a given video, $k$, there exist binary ground-truth labels $y^{(k)}_{t,c} \in \{0, 1\}$, where $t$ is the time-step and $c$ is the class. The network predicts class probabilities at each time-step, on which a threshold is applied to obtain binary predicted labels, $\tilde{y}^{(k)}_{t,c} \in \left\{0, 1\right\}$. Two standard metrics for multi-label classification are per-class precision and per-class recall, which are defined as:
\begin{equation}
    Precision(c) = \frac{ N_\text{correct}(c)}{ N_\text{predict}(c)}, Recall(c) = \frac{ N_\text{correct}(c)}{ N_\text{gt}(c)}.
\end{equation}
Here, $N_\text{correct}(c)=\sum_{k,t}{\mathbbm{1} [ y^{(k)}_{t,c} = \tilde{y}^{(k)}_{t,c} = 1 ]}$ are the number of correct predictions for class $c$,  $N_\text{predict}(c)=\sum_{k,t}{\mathbbm{1} [ \tilde{y}^{(k)}_{t,c} = 1 ]}$ are the total number of predictions for class $c$,  $N_\text{gt}(c)=\sum_{k,t}{\mathbbm{1} [ y^{(k)}_{t,c} = 1 ]}$ are the total number of time-steps containing class $c$, and $\mathbbm{1}$ is the indicator function. These metrics measure a model's performance on individual classes, but they do not take into account the relationships and dependencies between these classes. We propose action-conditional precision and recall to solve this issue.

We first deal with the co-occurrence relationship, where two actions occur within the same time-step. For an action class $c_i$, we measure its precision and recall when another action, $c_j$, is present within the same time-step. The action-conditional precision and recall of $c_i$, given $c_j$, are
\begin {eqnarray}
\label{eq:act-cond-cooccur}
Precision(c_i|c_j) &=& \frac{ N_\text{correct}(c_i|c_j)}{ N_\text{predict}(c_i|c_j)}, \text{ and} \nonumber \\
 Recall(c_i|c_j) &=& \frac{ N_\text{correct}(c_i|c_j)}{ N_\text{gt}(c_i|c_j)}.
\end {eqnarray}
Here, the components are defined as
\begin {eqnarray}
N_\text{correct}(c_i|c_j)&=&\sum_{k,t}{\mathbbm{1} [ y^{(k)}_{t,c_i} = \tilde{y}^{(k)}_{t,c_i} = 1 ] \mathbbm{1} [ y^{(k)}_{t,c_j} = 1 ]}, \nonumber \\
N_\text{predict}(c_i|c_j)&=&\sum_{k,t}{\mathbbm{1} [ \tilde{y}^{(k)}_{t,c_i} = 1 ] \mathbbm{1} [ y^{(k)}_{t,c_j} = 1 ]}, \text{ and}   \nonumber \\
N_\text{gt}(c_i|c_j)&=&\sum_{k,t}{\mathbbm{1} [ y^{(k)}_{t,c_i} = 1 ] \mathbbm{1} [ y^{(k)}_{t,c_j} = 1 ]} .
\end {eqnarray}
These metrics, measure the precision and recall of an action class $c_i$ when $c_j$ is present within the given time-step. Note that these metrics are not symmetric, and it may be the case that $Precision(c_i|c_j) \neq Precision(c_j|c_i)$ and $Recall(c_i|c_j) \neq Recall(c_j|c_i)$. 

These metrics measure co-occurrence within a time-step. We extend this to measure temporal dependencies between different actions, which follow each other within some temporal window $\tau$. We present metrics, which measure the precision and recall of action $c_i$, given that action $c_j$ was present within the last $\tau$ time-steps and $c_j$ is not present within the current time-step (this ensures that it measures only temporal dependencies and not co-occurrence dependencies). At time-step $t$, this holds when the following condition is true:
\begin{equation}
\label{eq:condition}
    y^{(k)}_{t,c_j} = 0 \land \exists y^{(k)}_{t^*,c_j} = 1, t^*\in[t-\tau,t).  
\end{equation}
Therefore, the action-conditional precision and recall, denoted $Precision(c_i|c_j,\tau)$ and $Recall(c_i|c_j,\tau)$, are computed with the following components:
\begin {eqnarray}
N_\text{correct}(c_i|c_j,\tau)&=&\sum_{k,t}{\mathbbm{1} [ y^{(k)}_{t,c_i} = \tilde{y}^{(k)}_{t,c_i} = 1 ] \mathbbm{1}[\chi]}, \nonumber\\
N_\text{predict}(c_i|c_j,\tau)&=&\sum_{k,t}{\mathbbm{1} [ \tilde{y}^{(k)}_{t,c_i} = 1 ] \mathbbm{1}[\chi]}, \text{ and} \nonumber\\
N_\text{gt}(c_i|c_j,\tau)&=&\sum_{k,t}{\mathbbm{1} [ y^{(k)}_{t,c_i} = 1 ] \mathbbm{1}[\chi]}.
\end {eqnarray}
Here, $\chi$ is the condition in equation \ref{eq:condition}. For ease of notation, we use $\tau=0$ to denote the action-conditional metrics within a time-step (equation \ref{eq:act-cond-cooccur}), such that $Precision(c_i|c_j,\tau=0) = Precision(c_i|c_j)$ and $Recall(c_i|c_j,\tau=0) = Recall(c_i|c_j)$.

Our proposed action-conditional metrics can be used to measure the co-occurrence dependencies and temporal dependencies between any two actions. Since some actions never co-occur or follow each other, the overall metric is computed by averaging all action pairs $(c_i,c_j), i\neq j,$ such that $N_\text{gt}(c_i|c_j,\tau)>0$. In addition, more complex performance metrics like F1-score (the harmonic mean between precision and recall) and mAP (the area under the precision-recall curve) can also be computed using our action-conditional precision and recall metrics.

\section{Experimental Evaluations}

\subsection{Experimental Setup}

\paragraph{Datasets}
We conduct experiments on two widely used multi-label action localization datasets: MultiTHUMOS \cite{yeung2018every} and Charades \cite{sigurdsson2016hollywood}. The MultiTHUMOS dataset is an extended version of THUMOS'14 \cite{THUMOS14} dataset, containing dense, multi-label frame-level action annotations for 65 classes across the 413 sports videos from YouTube. We use the standard train/test split with 200 videos for training and 213 for testing. MultiTHUMOS contains up to 25 action labels for each video, with an average of 10.5 activity instances per video and 1.5 labels per frame. This is in contrast to other activity detection datasets such as ActivityNet \cite{caba2015activitynet} and HACS \cite{zhao2019hacs}, which only have one activity per time-step. Charades\cite{sigurdsson2016hollywood} is a large dataset with 9848 videos of daily indoor activities, collected through Amazon Mechanical Turk. The dataset consists of 66,500 temporal annotations for 157 action classes. Contrary to MultiTHUMOS, the activities tend to be performed in the home. Each video in the dataset contains an average of 6.8 activity instances.

\paragraph{Implementation Details} In our experiments we use RGB and Optical Flow features extracted from two-stream I3D backbone pre-trained on Kinetics-400 dataset unless otherwise stated. A 1024 dimensional feature vector is extracted per stream from the final convolutional layer of an I3D \cite{carreira2017quo} network at 3 feature vectors per second from 24fps videos. Each feature vector corresponds to 8 frames or 0.33 seconds. The input sequence length is set to $T=128$ on MultiTHUMOS, and $T=64$ on Charades. For both datasets, our network uses $L=5$ MLAD layers (See section \ref{sec:ablations} for discussion on other values of $T$ and $L$). The dimension of the class-level feature vector, $H$, is set to 128 in all our experiments. During training, we classify and compute loss on both the initial class-level features and the features from the final MLAD layer (the effect of this loss computation is explored in the supplement). We train our models using Adam optimizer with an initial learning rate of 1e-4. All our models are trained on a single 32GB NVIDIA Tesla V100 GPU and implemented in PyTorch deep-learning framework.

\vspace{-0.3cm}
\paragraph{Baselines} We compare our method with several baselines. The first is a linear layer which classifies individual time-steps based on the features extracted from a pre-trained I3D network (denoted \textit{I3D Baseline}). A second baseline which extracts class-level features (as in equation \ref{eq:class-feats}) and classifies these features (as in equation \ref{eq:classifier-layer}) is also used (denoted \textit{CF Baseline}). In addition, we compare with recent multi-label action localization methods Super-events (SE) \cite{piergiovanni2018learning}, Temporal Gaussian Mixture (TGM) Layers \cite{piergiovanni2019temporal}, TGMs + SE \cite{piergiovanni2019temporal}, and TGMs + Differentiable Grammars (DG) \cite{piergiovanni2020differentiable}.

\vspace{-0.3cm}
\paragraph{Metrics} To compare with previous temporal action localization works, we use the standard evaluation protocol of computing per-frame mean average precision (f-mAP). We also present results on other multi-label metrics - Hamming Loss (HL), Zero One Loss (ZL), Ranking Loss (RL), Coverage Loss (CL), Jaccard Score (JS), and Label Ranking Average Precision (LRAP) - as well as our proposed action-conditional metrics: precision (${P}_{AC}$), recall (${R}_{AC}$), f1-score (${F1}_{AC}$), and mean average precision (${mAP}_{AC}$).

\begin{table}[t]
\centering
\begin{tabular}{l|c|c}
\hline
\textbf{Method}                                                & \textbf{MultiTHUMOS} & \textbf{Charades} \\ \hline
I3D Baseline* \cite{piergiovanni2019temporal}                         & 29.7  & 17.2    \\ 
CF Baseline                                                          & 42.6  & 14.8   \\
Super-events* \cite{piergiovanni2018learning}                         & 36.4  & 19.4   \\ 
TGMs* \cite{piergiovanni2019temporal}                                 & 44.3  & 21.5   \\ 
TGMs + SE*   \cite{piergiovanni2019temporal}                          & 46.4  & 22.3   \\ 
TGMs + DG* \cite{piergiovanni2020differentiable}                      & 48.2  & 22.9    \\ \hline
Our Approach                                                         & \textbf{51.5} & \textbf{23.7}    \\ \hline
\end{tabular}
\caption{Comparison of frame-level mAP score of our approach with previous works on MultiTHUMOS and Charades datasets using features from a pre-trained two-stream I3D model. Results indicated with * are from \cite{piergiovanni2020differentiable}.}
\label{tab:comparison}
\end{table}

\subsection{Results}

Our results on the MultiTHUMOS and Charades datasets are presented in Table \ref{tab:comparison}. Our approach achieves 51.5\% f-mAP and 23.7\% f-mAP on MultiTHUMOS and Charades respectively. The effectiveness of our MLAD layer is best illustrated by the comparison with CF Baseline: with only 5 MLAD layers, the class-based features are refined, leading to a 9\% improvement in f-mAP for both datasets.

\paragraph{Comparison with state-of-the-art}On MultiTHUMOS, our model outperforms the current state-of-the-art model, TGM + Differentiable Grammars, by 3.3\%; on Charades, we achieve a 0.8\% improvement in f-mAP. Although the absolute improvement is not as large as MultiTHUMOS (since it is a more difficult dataset with more action classes), the improvement is comparable to previous performance advancements on the dataset (e.g. 0.6\% improvement for TGM + DG over TGMs + SE).  


\vspace{-0.3cm}
\paragraph{Action-Conditional Metric Results}
We present results using other existing multi-label metrics (HL, ZL, RL, CL, JS, LRAP) alongside our proposed metrics (conditional precision, recall, f1-score, and mAP) on the MultiTHUMOS dataset in Table \ref{tab:metric}. For the time-conditional metrics, we select $\tau=20$; results with other values of $\tau$ are presented in the Supplementary Materials. Our method achieves higher performance on all action-conditional metrics since it  models different action dependencies within a video, both within a time-step ($\tau=0$) and throughout time ($\tau>0$). Of the 1322 action pairs that co-occur within a time-step in the test set, our method improves the average precision of 961 pairs when compared to the I3D baseline. 

By analyzing specific action pairs, one can better understand how various approaches model the different action dependencies. Here, we examine the dependencies described in Figure \ref{fig:motivation}. To evaluate how well the method models the co-occurrence dependency between actions ``Basketball Dribble" and ``Run", one can compute the average precision over that pair: $AP(c_i=\text{BasketballDribble}|c_j=\text{Run}, \tau=0)$. The TGM approach achieves a minor improvement over the I3D baseline (47.26\% vs 45.60\%), while our approach better models this relationship with an average precision of 58.85\%. A similar improvement is seen for temporal dependencies. To evaluate the relationship ``Fall follows Jump", we compute $AP(c_i=\text{Fall}|c_j=\text{Jump}, \tau=20)$, and find that our method achieves a score of $78.12\%$ compared to the TGM's $72.27\%$.

\begin{table*}[t]
\resizebox{\textwidth}{!}{%
\centering
\begin{tabular}{l|c|c|c|c|c|c|c|c|c|c|c|c|c|c}
& \multicolumn{5}{c|}{\textbf{Existing Metrics}} & \multicolumn{8}{c}{\textbf{Action-Conditional Metrics} $\uparrow$} \\ 
\hline
 &
\textbf{HL $\downarrow$} & \textbf{ZL $\downarrow$} & \textbf{RL $\downarrow$} & \textbf{CE $\downarrow$} & \textbf{JS $\uparrow$} & \textbf{LRAP $\uparrow$} & \multicolumn{4}{c|}{\textbf{$\tau=0$}} & \multicolumn{4}{c}{\textbf{$\tau=20$}} \\ \cline{8-15} 
 &  &  &  &  &  &  & 
$\textbf{P}_{AC}$ & $\textbf{R}_{AC}$ & $\textbf{F1}_{AC}$ & $\textbf{mAP}_{AC}$ & $\textbf{P}_{AC}$ & $\textbf{R}_{AC}$ & $\textbf{F1}_{AC}$ & $\textbf{mAP}_{AC}$ \\ \hline
    I3D & 0.018 & 0.673 & 0.029 & 4.409 & 0.260 & 0.770 & 33.63 & 15.23 & 18.65 & 32.58 & 37.88 & 18.01 & 21.96 & 35.53 \\ 
    CF & 0.017 & 0.646 & 0.027 & 4.242 & 0.315 & 0.787 & 36.73 & 21.39 & 23.71 & 35.00 & 41.95 & 23.91 & 27.22 & 38.42 \\ 
    TGM \cite{piergiovanni2019temporal} & 0.017 & 0.642 & 0.022 & 3.798 & 0.297 & 0.800 & 34.59 & 17.21 & 20.14 & 36.90 & 39.27 & 20.13 & 23.86 & 40.18 \\ 
    Our & \textbf{0.017} & \textbf{0.635} & \textbf{0.017} & \textbf{3.276} & \textbf{0.373} & \textbf{0.816} & \textbf{39.22} & \textbf{28.33} & \textbf{29.37} & \textbf{40.15} & \textbf{42.89} & \textbf{30.27} & \textbf{32.18} & \textbf{43.76} \\ 
    \hline
\end{tabular}}
\caption{Evaluation of our approach using existing multi-label classification metrics and our proposed action dependency metrics on MultiTHUMOS dataset. \textbf{HL} - Hamming Loss, \textbf{ZL} - Zero One Loss, \textbf{RL} - Ranking Loss, \textbf{CL} - Coverage Loss, \textbf{JS} - Jaccard Score, \textbf{LRAP} - Label Ranking Average Precision, $\textbf{P}_{AC}$ - Action-Conditional Precision, $\textbf{R}_{AC}$ - Action-Conditional Recall, $\textbf{F1}_{AC}$ - Action-Conditional F1-Score, $\textbf{mAP}_{AC}$ - Action-Conditional Mean Average Precision.
}
\label{tab:metric}
\end{table*}

\begin{table}[t]
\centering
\begin{tabular}{l|c|c}
\hline
\textbf{}       & \textbf{MultiTHUMOS} & \textbf{Charades} \\ \hline
\textbf{L = 1}  & 48.55        & 20.48     \\ \hline
\textbf{L = 3}  & 50.30        & 23.15    \\ \hline
\textbf{L = 5}  & 51.52        & 23.74   \\ \hline
\end{tabular}
\caption{Ablation on the number of MLAD layers, $L$. As the number of layers increases, we observe an increase in performance with diminishing gain. f-mAP is the evaluation metric used.}
\label{tab:abl:nlayersthumos}
\vspace{-0.3cm}
\end{table}

\subsection{Ablations}
\label{sec:ablations}
We evaluate the various design decisions for our method as well as its components.

\vspace{-0.3cm}
\paragraph{Number of MLAD Layers} Since our proposed MLAD layer can be stacked to continually refine input features, we test how performance changes as the number of MLAD layers increases. We show in Table \ref{tab:abl:nlayersthumos} that increasing the number of layers tends to improve results on both MultiTHUMOS and Charades. However, this improvement has diminishing gains as the depth increases: the change from 3 to 5 layers leads to a smaller improvement (1.22\% on MultiTHUMOS) than the change from 1 to 3 layers (1.85\%). An increase to $L=7$ leads to no noticeable improvement, therefore all reported results have a depth of $L=5$.

\vspace{-0.3cm}
\paragraph{Effect of Feature Sequence Length} Since our approach performs computations on a feature sequence of length $T$, we evaluate how the sequence length affects our performance. We present two experimental setups: 1) both the training and evaluation lengths are fixed, and 2) the training length is varied, $T \in \{i\times16 \mid i \in \{1, ... ,8\}\}$, with a fixed evaluation length. 
We present the results of both in Table \ref{tab:abl:vartime}.
We find that when the training length is fixed, the performance peaks when $T=96$ at $51.31\%$ mAP. However, when the training length is varied in experiment setup 2, we achieve the best performance with $T=128$. This varying of sequence length during training can be seen as a form of data augmentation, leading to improved generalization.

\begin{table}[t]
\centering
\begin{tabular}{l|c|c}
\hline
\textbf{Eval. Length}  & \textbf{Fixed Tr. Length} & \textbf{Var. Tr. Length}\\ \hline
\textbf{T = 32}                                  & 49.90  &   50.20     \\ \hline
\textbf{T = 64}                                  & 51.14  &   51.01     \\ \hline
\textbf{T = 96}                                  & 51.31  &   51.31     \\ \hline
\textbf{T = 128}                                 & 50.59  &   51.52     \\ \hline
\end{tabular}
\caption{Ablation for fixed and variable length training sequences, with fixed evaluation lengths. f-mAP is the evaluation metric used.}
\label{tab:abl:vartime}
\end{table}

\begin{table}[t]
\centering
\begin{tabular}{l|c|c}
\hline
\textbf{}          & \textbf{MultiTHUMOS} & \textbf{Charades} \\ \hline

\textbf{No CB, No TB}  &        42.60      & 14.80                     \\ \hline
\textbf{Only CB}  &        44.98      & 20.3                      \\ \hline
\textbf{Only TB}  &       48.03       & 21.1                      \\ \hline
\textbf{TB + CB} & 51.52  & 23.5 \\ \hline
\end{tabular}
\caption{Ablation on the two branches (CB and TB) of the MLAD layer. f-mAP is the metric reported here.
}
\label{tab:abl:branch}
\end{table}

\vspace{-0.3cm}
\paragraph{Effect of CB and TB} Since both branches of the MLAD layer are meant to model the different action dependencies within a video, we run an ablation by removing each of these branches and present the results in Table \ref{tab:abl:branch}. We find that each branch leads to improvement over classifying with the original class-level features, but that best performance is achieved when both are used.

\begin{figure}[t]
\begin{center}
\includegraphics[width=\linewidth]{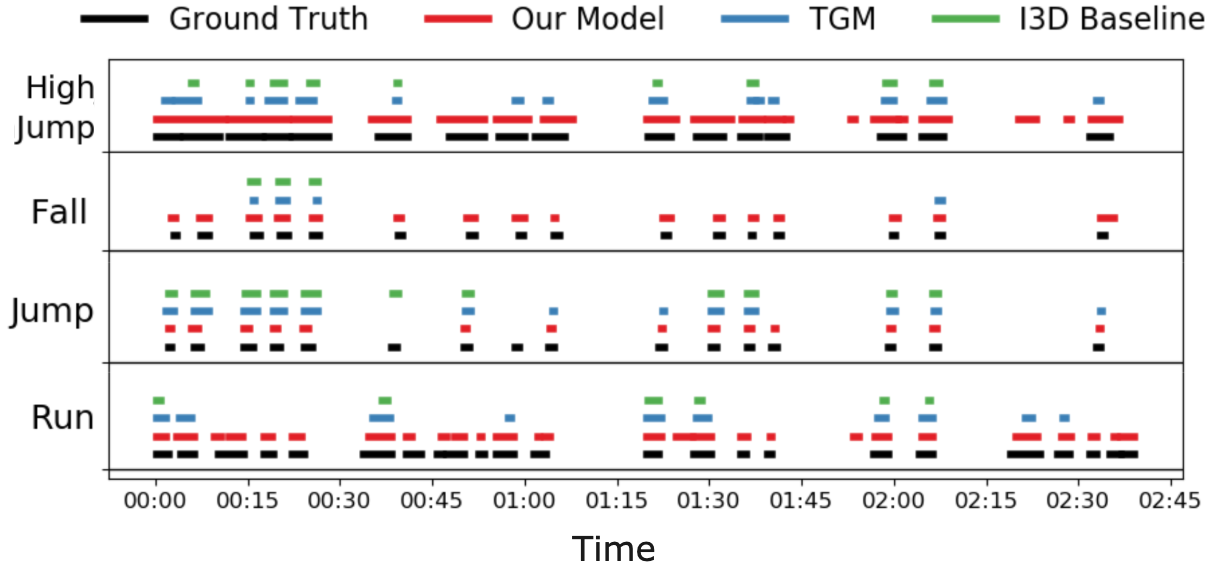}
\end{center}
\caption{Visualization of model predictions for different time steps (x-axis) for various actions (y-axis) from a MultiTHUMOS test video. We compare our model (red), with TGM \cite{piergiovanni2019temporal} (blue), I3D baseline (green), and the ground-truth (black).
The performance of our model in detecting temporally dependent actions  ``Jump" and ``Fall" is higher than the baselines; between time-steps 00:30 to 02:00 our method detects all instances of ``Fall" while TGM and the I3D baseline have no predictions for this class.   
}
\label{fig:predictions}
\end{figure}

\section{Discussion and Analysis} 

In this section, we analyze our trained model's predictions and the learned attention maps from MLAD layers.

\vspace{-0.4cm}
\paragraph{Localizaton Analysis}
In Figure \ref{fig:predictions}, we visualize the predictions of our trained model on a sample video sequence from the MultiTHUMOS test set. 
When compared with the outputs from the TGM \cite{piergiovanni2019temporal} network and the I3D baseline, our proposed method generates localizations that better overlap with the ground-truth annotations. This is most notable for the ``Fall" action; the MLAD layers allow our method to model the temporal dependencies between ``Fall" and ``Jump", leading to the improved localization predictions.  
Also, our model detects every instance of ``HighJump" which co-occurs with other actions ``Run", ``Jump", and ``Fall". In general, we find that our approach leads to a large increase in recall across most classes: of the 65 actions in the MultiTHUMOS dataset, our network improves recall for 52 classes, with an average improvement of 13.17\%. Additional analysis of the results, including per-class scores, can be found in the Supplementary Materials.

\begin{figure}[t!]
\begin{center}
\includegraphics[width=0.97\linewidth]{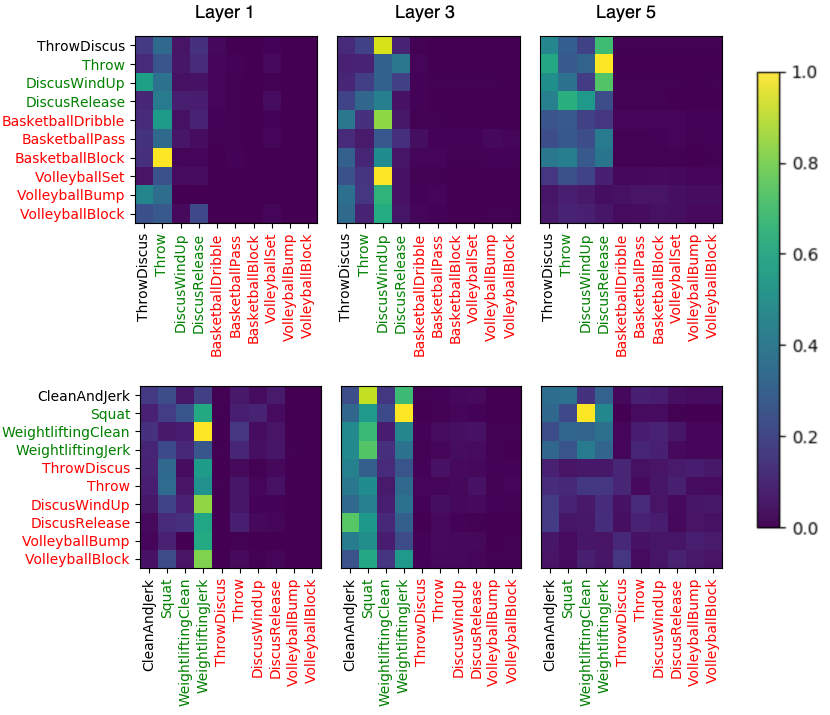}
\end{center}
\caption{Visualization of $10\times10$ subsets of the CB attention maps from MLAD layers 1, 3, and 5 obtained by averaging the maps over time-steps where action ``Throw Discus" (top) and ``Clean and Jerk" (bottom) are present. Columns corresponding to classes that are related to the main action (those in green) tend to be active, whereas unrelated classes (in red) tend to have low activation.
} 
\label{fig:class_attention2_individual}
\vspace{-0.3cm}
\end{figure}

\vspace{-0.3cm}
\paragraph{Failure Cases}
When compared to previous approaches, our network tends to under-perform on the ``Walk" and ``Sit" actions on the MultiTHUMOS dataset. We find that these actions tend to occur in the background (e.g. by a referee or audience members at a sporting event) and frequently co-occur with many different foreground actions. Since these background actions are not directly dependent on those in the foreground, our method attempts to model relationships that do not exist, leading to poor performance. This suggests that the modeling of individual actors would be greatly beneficial for learning the various action dependencies within a video. We believe that this would be a promising direction for future work.




\vspace{-0.3cm}
\paragraph{Interpretability of MLAD Layers}
One advantage of our approach over previous temporal action localization methods is that the network architecture (specifically the MLAD layers) allows for more interpretable results. Since the MLAD Layers consist of two attention-based branches - the Co-occurrence Dependency Branch (CB) and the Temporal Dependency Branch (TB) - we can analyze their attention maps to better understand how the different action dependencies are modeled. These analyses are done on the MultiTHUMOS dataset where there are 65 action classes ($C=65$) and a sequence length of 128 ($T=128$) is used. 

\begin{figure}[t!]
\begin{center}
\includegraphics[width=1.0\linewidth]{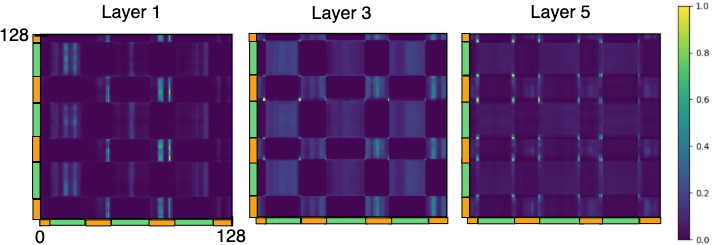}
\end{center}
\caption{Visualization of the $T \times T$ ($T$ = 128) TB attention maps from  MLAD layers 1, 3 and 5 on a sample sequence for the action ``Close Up Talk To Camera". The checker-board pattern shows that time-steps where the action is present (highlighted in green) tend to focus on other time-steps where it is present; and time-steps where the action is absent (in orange) 
tend to focus on other time-steps where it is absent. We also find that various layers model different temporal regions (e.g. layer 5 models the boundaries of the action for class ``Close Up Talk To Camera"). 
} 
\label{fig:time_attention_individual}
\vspace{-0.3cm}
\end{figure}

We first analyze attention maps in the CB. Figure \ref{fig:class_attention2_individual} contains $10\times10$ subsets of the $C \times C$ attention maps from different MLAD layers, which are obtained by averaging over all time-steps where a specific action is present (``Throw Discus" on the top and ``Clean and Jerk" on the bottom). We find that this model successfully models the co-occurrence dependencies since the actions which are related to the present action (e.g. ``Throw", ``Discus Wind-Up", and ``Discuss Release" often co-occur with ``Throw Discus") tend to be active, whereas unrelated actions (e.g. basketball and volleyball actions) tend to have low activation. We also find that the CB focuses on actions, like ``Run" and ``Jump", which are most prevalent in the training set\footnote{Additional visualizations which illustrate this behavior can be found in the Supplementary Material.} - this is likely because these actions often co-occur with many different actions, so their presence (or absence) is important in determining the existence of other less common actions.

Next, we present the attention maps from the TB in Figure \ref{fig:time_attention_individual}. We visualize the $T \times T$ maps for the class ``Close Up Talk To Camera" from a sample sequence. A noticeable checker-board pattern is present: Time-steps, where the action is present, tend to focus on other time-steps where it is present, while time-steps, where the action is absent, tend to focus on other time-steps where it is absent. This behavior is common across all actions. Furthermore, we find that different MLAD layers attend to different parts of an action; for example, in layer 5 the attention map is active at the action boundaries for the ``Close Up Talk To Camera" action. We provide attention maps for all MLAD layers, as well as more examples, in the supplement.

\section{Conclusion}
In this work, we propose an attention-based network architecture to learn action dependencies in videos, for solving the multi-label temporal action localization task. Our proposed MLAD layer consisting of two branches: The co-occurrence Dependency Branch and the Temporal Dependency Branch, which use attention to model dependencies between actions that occur within the same time-step, and those actions which precede/follow each other, respectively. 
As the existing evaluation metrics for multi-label temporal localization do not explicitly consider action dependencies, we propose a novel evaluation metric. Our method out-performs the current state-of-the-art on existing multi-label classification metrics as well as our proposed metric. 

\paragraph{Acknowledgments}
This research is based upon work supported by the Office of the Director of National Intelligence(ODNI), Intelligence Advanced Research Projects Activity (IARPA), via IARPA R\&D Contract No. D17PC00345. The views and conclusions contained herein are those of the authors and should
not be interpreted as necessarily representing the official policies or endorsements, either expressed or implied, of the ODNI, IARPA, or the U.S. Government. The U.S. Government is authorized to reproduce and distribute reprints for Governmental purposes not withstanding any copyright annotation thereon.

\clearpage

\setcounter{equation}{0}
\setcounter{figure}{0}
\setcounter{table}{0}
\setcounter{section}{0}
\makeatletter
\renewcommand{\theequation}{S\arabic{equation}}
\renewcommand{\thefigure}{S\arabic{figure}}
\renewcommand{\thetable}{S\arabic{table}}




\begin{center}
\Large{\textbf{Supplementary}}
\end{center}

\section{Overview}
In this supplement, we present additional ablation experiments (Section \ref{ablation}) and a more detailed comparison with other methods (Section \ref{comparison}). We also present additional information about the parameters in our MLAD layer (Section \ref{parameters}) and experiments on an additional baseline (Section \ref{baseline}). Lastly, we present additional results using our proposed metric (Section \ref{metric}) and additional visualizations of learned attention maps (Section \ref{analysis}).


\section{Ablation Experiments} \label{ablation}

\paragraph{Effect of RGB and Flow} We analyse how the use of features from different modalities (RGB and optical flow) effect our method in Table \ref{tab:abl:rgbflow}. Consistent with previous action localization works \cite{piergiovanni2019temporal} we find that the use of Flow features outperforms using only RGB (by a margin of 6.53\% on MultiTHUMOS and 1.7\% on Charades), but the combination of both features leads to the best results. {\em Notably, our method only trained using optical flow features outperforms all previous methods, even when they use both RGB and Flow features. }

\begin{table}[h]
\centering
\begin{tabular}{l|c|c}
\hline
\multicolumn{1}{c|}{\textbf{Features}} & \textbf{MultiTHUMOS} & \textbf{Charades}       \\ \hline
\textbf{RGB}                           & 42.24                 & 18.40                   \\ \hline
\textbf{Flow}                          & 48.77                 & 20.10                   \\ \hline
\textbf{Late Fusion}                   & 49.58                 & 22.93                   \\ \hline
\textbf{Early Fusion}                  & 51.52                 & 23.74                   \\ \hline
\end{tabular}
\caption{Ablation results using RGB, Flow and their combination. f-mAP is the evaluation metric. In Early Fusion, the network is trained with concatenated RGB and Flow features. In Late Fusion, two networks are trained with RGB, Flow features independently and their predictions are averaged.}
\label{tab:abl:rgbflow}
\end{table}

\begin{table}[t]
\centering
\begin{tabular}{l|c|c}
\hline
\textbf{} & \textbf{Fixed Alpha} & \textbf{Learned Alpha} \\ \hline
\textbf{f-mAP} & 50.95 & 51.52 \\ \hline
\end{tabular}
\caption{Ablation results showing the effect of weight parameter $\alpha$. Learning the $\alpha$ parameter gives better results when compared to fixed $\alpha$ = 0.5.}
\label{tab:abl_alpha}
\end{table}

\paragraph{Effect of Learned Feature Averaging} When combining the refined features from the Co-occurrence Dependency Branch and the Temporal Dependency Branch, we use a weighted average based on a learned parameter $\alpha$. We compare the performance of our model trained with additional $\alpha$ parameter and the model trained with a fixed value $\alpha$ = 0.5. As shown in Table \ref{tab:abl_alpha}, learning the weight parameter $\alpha$ gives better results when compared to averaging ($\alpha$ = 0.5) the features from TB and CB branches. Although the absolute difference in score is not large, this 0.57\% improvement in f-mAP is the result of only 5 learned parameters (since there are 5 MLAD layers). 

\begin{table}[]
\centering
\begin{tabular}{l|c|c}
\hline
\textbf{} & \textbf{W/O Initial Loss} & \textbf{With Initial Loss} \\ \hline
\textbf{f-mAP} & 49.96 & 51.52 \\ \hline
\end{tabular}
\caption{Ablation results with and without classification loss on the class-specific features.}
\label{tab:abl_init_classification}
\end{table}

\paragraph{Effect of initial Classification Loss} Classification step in our proposed model consists of two parts: (i) classification of initial class-specific features before the MLAD layers. (ii) classification of refined features after the MLAD layers. In Table \ref{tab:abl_init_classification} we present the results with and without the classification loss on the initial class-specific features. We find that the use of classification loss on the initial class-specific features increases performance since it gives additional supervisory signal which improves the class-level representations given to the MLAD layers.

\begin{figure}[t]
\begin{center}
\includegraphics[width=\linewidth]{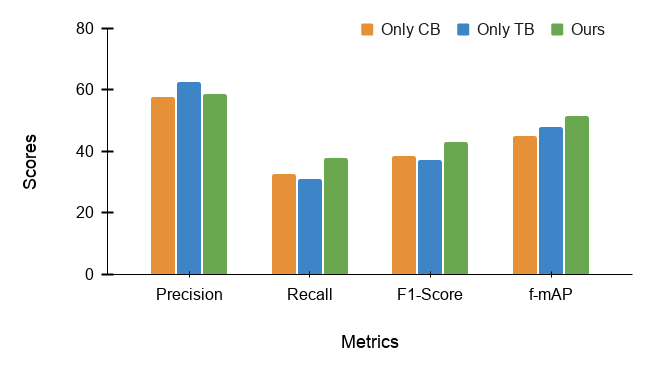}
\end{center}
\caption{Comparison of performance metrics between different variants of our proposed model. ``Only CB" refers to model trained with Co-occurrence Dependency Branch (CB), ``Only TB" refers to model trained with Temporal Dependency Branch (TB), and ``Ours" refers to model trained with both the branches.}
\label{fig:cb_tb}
\end{figure}

\paragraph{Effect of CB and TB} In Figure \ref{fig:cb_tb} we compare the mean Precision, Recall, F1-Score and f-mAP over all the classes in MultiTHUMOS dataset for three different models. ``Only CB" refers to the model with only the Co-occurrence Dependency Branch in a MLAD layer, ``Only TB" refers to the model with only the Temporal Dependency Branch and ``Ours" refers to the model containing both the branches. 
We observe that Co-occurrence Dependency Branch helps in detecting instances of actions which co-occur within a time-step and thereby improve the overall recall of the model. The Temporal Dependency Branch, on the other hand, improves the temporal boundaries of action instances and tends to increase precision. 

\section{Class-level Performance} \label{comparison}

\begin{figure*}[t!]
\begin{center}
\includegraphics[width=\linewidth]{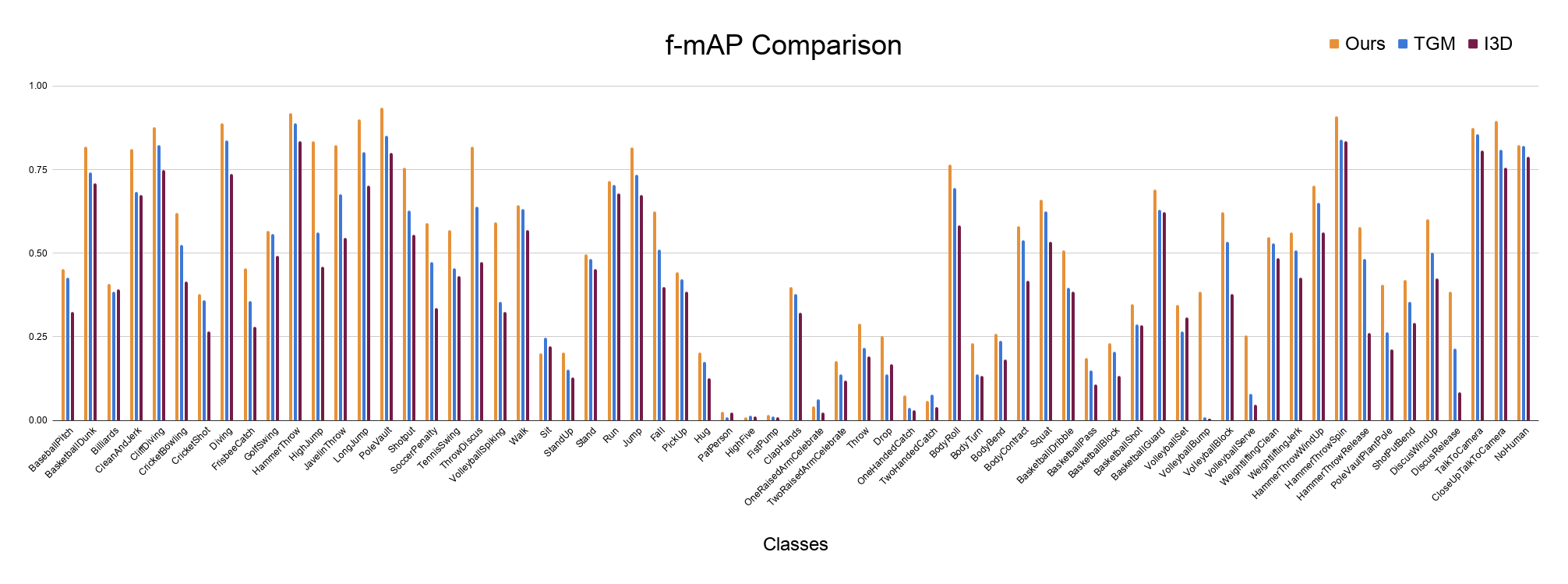}
\end{center}
\caption{Per-class f-mAP comparison between our model, TGM\cite{piergiovanni2019temporal} and I3D\cite{carreira2017quo}.} 
\label{fig:per-class}
\end{figure*}

In Figure \ref{fig:per-class}, we present the per-class f-mAP score comparison between the I3D baseline, TGM \cite{piergiovanni2019temporal},
and our proposed model on MultiTHUMOS dataset. Please refer to provided videos for a further comparison between TGM and our model. Our model outperforms previous methods in most of the classes, with one notable except ``Sit". We observe that action ``Sit" is associated with a background actor and is not related to the foreground action. Our proposed model, under-performs when compared to TGM in modeling the relationship between foreground action and this unrelated background action. However, our method does successfully model relationships between actions performed by the actor in the foreground. This is evident in the provided video, ``video\_test\_0000444.avi". 

\section{Parameters}
\label{parameters}

Each MLAD layer contain about 200K parameters and can be easily stacked on top of any existing feature extraction backbone with a minor change to learn the action dependencies for end-to-end action classification/localization. The output features from the feature extractor need to be class specific before they can be processed by a MLAD layer. Number of parameters in a MLAD layer is insignificant when compared to the number of parameters in a typical backbone, two-stream I3D contain 25M parameters, so adding MLAD layers will not contribute to a huge increase in parameters and thereby memory consumption.

\section{Baseline - Fully Connected Attention}\label{baseline}

The trivial approach to learn relationships between actions across time-steps would be to use a fully-connected attention-based network to learn all $CT\times CT$ relationships, where $C$ is the number of classes and $T$ is the number of time-steps in the input. To evaluate the efficiency of this approach, we trained a network where we replace the Co-occurrence Dependency Branch (CB) and Temporal Dependency Branch (TB) in our model with a self-attention module modeling the $CT\times CT$ relationships. 

As the memory consumption of this models increases quadratically with increase in number of classes ($C$) or the number of timesteps ($T$), it is infeasible to train when either of these become too large. We find that on a 32GB GPU, we must reduce the batch size for both datasets to 8 (batch size = 32 is used to train our model). Furthermore, the maximum number of layers which can fit into memory are $L=3$ for the MultiTHUMOS dataset with 65 classes and $L=1$ for the Charades dataset with 157 classes. 

In Table \ref{tab:baseline}, we compare the results of our model with this baseline. We observe that this trivial attention baseline with $CT\times CT$ connections performs worse than our proposed method which decomposes the attention operation into $C\times C$ and $T\times T$ sets of connections. Furthermore, we find that increasing the number of layers in this baseline does not lead to an substantial increase in performance (only a difference of 0.11\% when increasing from 1 to 3 layers). This suggests that even if more memory was available (i.e. through multiple GPUs or improvements in GPU technology), it is unlikely that it would be able to outperform our proposed MLAD layer.



\begin{table}[]
\centering
\resizebox{\columnwidth}{!}{
\begin{tabular}{l|c|c|c|c}
\multirow{2}{*}{} & \multicolumn{2}{c|}{\textbf{MultiTHUMOS}} & \multicolumn{2}{c}{\textbf{Charades}} \\ \cline{2-5} 
 & FC Baseline & \multicolumn{1}{l|}{Our Model} & \multicolumn{1}{l|}{FC Baseline} & \multicolumn{1}{l}{Our Model} \\ \hline
\textbf{L = 1} & 48.10 & 48.56 & 20.72 & 20.48 \\ \hline
\textbf{L = 3} & 48.21 & 50.30 & - & 23.15 \\ \hline
\textbf{L = 5} & - & 51.52 & - & 23.74 \\ \hline
\end{tabular}
}
\caption{Comparison of our results with the baseline model containing a self-attention layer modeling relationships between all the classes and timesteps. f-mAP is the evaluation metric used and the baseline results are presented only for those settings which fit in 32GB GPU used for training other models.}
\label{tab:baseline}
\end{table}

\section{Proposed Metric} \label{metric}
We present additional results for our proposed metric on MultiTHUMOS with various values of $\tau>0$ in Table \ref{tab:metric1}. By varying the $\tau$ parameter, we can measure short-term or long-term temporal dependencies. In general, we find that our proposed method outperforms the previous approaches on all metrics. We also evaluate our method, and our baselines, on the Charades dataset with the action-conditional metric. These results are shown in Table \ref{tab:metric2}.

As stated in the main text, one benefit of our proposed action-conditional metric is that we can obtain the pair-wise performance between two dependant action classes. We present the pairwise scores for various action classes in the MultiTHUMOS dataset with co-occurrence dependancies ($\tau=0$) and temporal dependencies ($\tau=20$) in tables \ref{tab:pairwise1} and \ref{tab:pairwise2} respectively. These results show that our method successfully models the various dependencies between different actions. Furthermore, it suggests that improvements in previous methods for temporal action localization do not necessarily come from improved modeling of action dependencies. Although TGM greatly outperforms the I3D baseline on this dataset, there are several action relationships on which it performs worse than the baseline (e.g. ``CricketBowling" and ``Throw" in table \ref{tab:pairwise1}, and ``Volleyball Spiking" and ``VolleyballSet" in table \ref{tab:pairwise2}). 

\begin{table*}[t]
\resizebox{\textwidth}{!}{
\centering
\begin{tabular}{l|cccc|cccc|cccc}
  & \multicolumn{4}{c|}{\textbf{$\tau=5$}} & \multicolumn{4}{c|}{\textbf{$\tau=10$}} & \multicolumn{4}{c}{\textbf{$\tau=40$}} \\ 
\hline

 & $\textbf{P}_{AC}$ & $\textbf{R}_{AC}$ & $\textbf{F1}_{AC}$ & $\textbf{mAP}_{AC}$ & $\textbf{P}_{AC}$ & $\textbf{R}_{AC}$ & $\textbf{F1}_{AC}$ & $\textbf{mAP}_{AC}$ & $\textbf{P}_{AC}$ & $\textbf{R}_{AC}$ & $\textbf{F1}_{AC}$ & $\textbf{mAP}_{AC}$ \\ \hline
    I3D & 33.23 & 17.87 & 20.47 & 36.78 & 36.19 & 18.17 & 21.65 & 36.26 & 39.27 & 17.71 & 21.98 & 34.74 \\ 
    CF & 36.96 & 23.43 & 25.08 & 39.07 & 40.19 & 24.04 & 26.79 & 39.09 & 42.67 & 23.44 & 27.04 & 37.58 \\ 
    TGM \cite{piergiovanni2019temporal} & 34.48 & 19.22 & 21.54 & 40.50 & 37.71 & 20.35 & 23.56 & 40.70 & 40.85 & 20.08 & 24.21 & 39.68 \\ 
    Ours & \textbf{40.13} & \textbf{29.60} & \textbf{30.16} & \textbf{43.88} & \textbf{42.85} & \textbf{30.67} & \textbf{32.25} & \textbf{44.43} & \textbf{44.01} & \textbf{30.25} & \textbf{32.53} & \textbf{43.34} \\ \hline
\end{tabular}}
\caption{Evaluation of various methods on the MultiTHUMOS dataset using our proposed action-conditional metric with varying values of $\tau$. $\textbf{P}_{AC}$ - Action-Conditional Precision, $\textbf{R}_{AC}$ - Action-Conditional Recall, $\textbf{F1}_{AC}$ - Action-Conditional F1-Score, $\textbf{mAP}_{AC}$ - Action-Conditional Mean Average Precision.
}
\label{tab:metric1}
\end{table*}

\begin{table*}[t]
\centering
\begin{tabular}{l|cccc|cccc|cccc}
  & \multicolumn{4}{c|}{\textbf{$\tau=0$}} & \multicolumn{4}{c|}{\textbf{$\tau=20$}} & \multicolumn{4}{c}{\textbf{$\tau=40$}} \\ 
\hline

 & $\textbf{P}_{AC}$ & $\textbf{R}_{AC}$ & $\textbf{F1}_{AC}$ & $\textbf{mAP}_{AC}$ & $\textbf{P}_{AC}$ & $\textbf{R}_{AC}$ & $\textbf{F1}_{AC}$ & $\textbf{mAP}_{AC}$  & $\textbf{P}_{AC}$ & $\textbf{R}_{AC}$ & $\textbf{F1}_{AC}$ & $\textbf{mAP}_{AC}$\\ \hline
    I3D & 14.34 & 1.33 & 2.10 & 15.17 & 12.68 & 1.94 & 2.93 & 21.43 & 14.93 & 2.02 & 3.07 & 20.26 \\ 
    CF & 10.27 & 1.04 & 1.63 & 15.77 & 9.01 & 1.50 & 2.23 & 22.23 & 10.69 & 1.58 & 2.36 & 21.04 \\ 
    Ours & \textbf{19.33} & \textbf{7.23} & \textbf{8.86} & \textbf{28.94} & \textbf{18.85} & \textbf{8.88} & \textbf{10.52} & \textbf{35.74} & \textbf{19.64} & \textbf{9.04} & \textbf{10.77} & \textbf{ 34.78}\\ \hline
\end{tabular}
\caption{Evaluation of various methods on the Charades dataset using our proposed action-conditional metric. $\textbf{P}_{AC}$ - Action-Conditional Precision, $\textbf{R}_{AC}$ - Action-Conditional Recall, $\textbf{F1}_{AC}$ - Action-Conditional F1-Score, $\textbf{mAP}_{AC}$ - Action-Conditional Mean Average Precision.
}
\label{tab:metric2}
\end{table*}

\begin{table*}[t]
\centering
\begin{tabular}{l|l|cccc}
  Action1 & Action2 & I3D Baseline & CF Baseline & TGM \cite{piergiovanni2019temporal} & Ours \\ 
\hline
BasketballDunk &    Jump    & 88.12 & 91.32 & 90.46 & \textbf{93.18} \\
CliffDiving    &      Jump   &              86.18 & 88.20 & 90.86 & \textbf{95.01}\\
VolleyballSpiking &   Jump   &              45.97 & 52.32 & 50.77 & \textbf{64.63}\\
PickUp      &         Squat  &              62.34 & 63.17 & 60.79 & \textbf{65.29}\\
Throw        &        HammerThrow &         29.89 & 40.57 & 36.74 & \textbf{47.49}\\
BodyBend     &        Diving      &         36.89 & 39.73 & 42.28 & \textbf{45.45}\\
BasketballDribble &   Run       &       45.60 & 55.33 & 47.26 & \textbf{58.85}\\ 
CricketBowling   &    Throw     &          78.23 & 78.18 & 77.83 & \textbf{87.16}\\ 

 \hline
\end{tabular}
\caption{Action-conditional scores for some actions pairs with co-occurrence dependencies in the MultiTHUMOS dataset. The metric reported is is aciton-conditional average precision for \textit{Action1} given \textit{Action2}: AP$(i=\text{Action1}|j=\text{Action2}, \tau=0)$.}
\label{tab:pairwise1}
\end{table*}

\begin{table*}[t]
\centering
\begin{tabular}{l|l|cccc}
  Action1 & Action2 & I3D Baseline & CF Baseline & TGM \cite{piergiovanni2019temporal} & Ours \\ 
\hline
BasketballDunk     &   BasketballShot       & 67.14 & 73.00 & 67.96 & \textbf{74.60} \\
VolleyballSpiking  &   VolleyballSet        & 60.99 & 66.22 & 51.14 & \textbf{70.21} \\
Fall               &   Jump                 & 63.38 & 72.96 & 72.27 & \textbf{78.12} \\
FrisbeeCatch       &   Throw                & 37.33 & 39.51 & 43.30 & \textbf{59.22} \\
BasketballGuard    &   BasketballShot       & 67.11 & 69.35 & 66.21 & \textbf{75.95} \\
HammerThrowSpin    &   HammerThrowWindUp    & 96.52 & 97.32 & 96.28 & \textbf{97.47} \\
HammerThrowRelease &   HammerThrowSpin      & 61.52 & 69.46 & 72.89 & \textbf{75.35} \\
DiscusRelease      &   DiscusWindUp         & 65.88 & 65.55 & 63.41 & \textbf{69.51} \\ 

 \hline
\end{tabular}
\caption{Action-conditional scores for some actions pairs with temporal dependencies in the MultiTHUMOS dataset. The metric reported is is action-conditional average precision for \textit{Action1} given \textit{Action2}: AP$(i=\text{Action1}|j=\text{Action2}, \tau=20)$.}
\label{tab:pairwise2}
\end{table*}


\section{Analysis} \label{analysis}
\begin{figure*}[ht!]
\begin{center}
\includegraphics[scale=0.4]{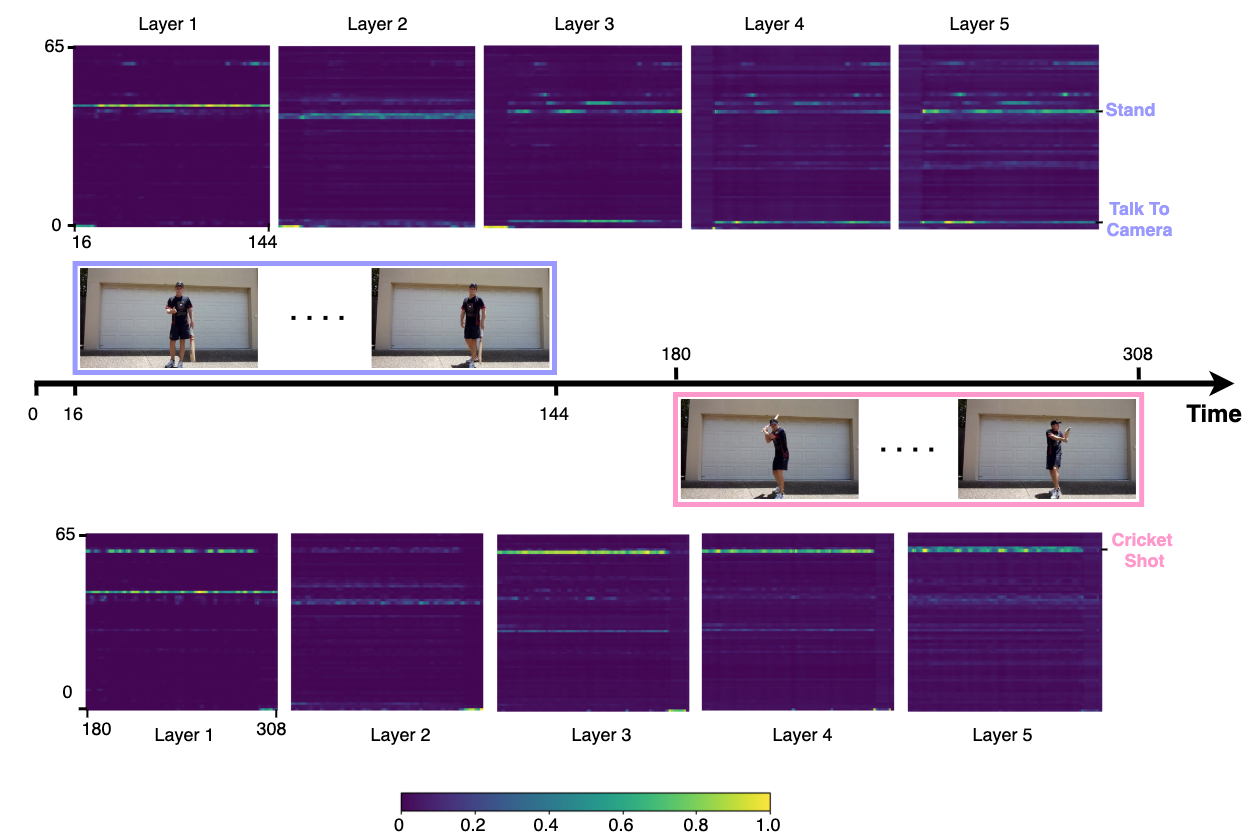}
\caption{Visualization of class attention maps from every MLAD layer of our network for two sequences from the same video. Top row is for the sequence on the left in the middle row, and the bottom row is for the sequence on the right. Attention maps shown here are of dimension $C \times T$ ($T$ = 128 (time steps), $C$ = 65 (classes)) and are obtained by averaging the $T$ class attention maps, with dimensions $C \times C$, along the columns. This gives the average attention for a given class at each time-step. Labels of the action classes (Stand, Talk to Camera, Cricket Shot) present in these selected sequences are shown on the right at their corresponding indices. Start and end index for the sequence is shown on the x-axis. 
} 
\end{center}
\label{fig:class_attention_individual}
\end{figure*}

In Figure \ref{fig:class_attention_individual}, we present  attention maps averaged over each column of the $C \times C$ map. The resultant $C \times T$ attention map shows the average attention received by a class from other classes for each of the time-steps in the input sequence. The top and bottom row in this figure show the $C \times T$ attention maps for two sequences from a sample video, shown on the left and the right in the middle row. Attention maps from the initial layers show activation for the frequent classes in the dataset which co-occur with most of the other classes, where as active regions in the attention maps from deeper layers correspond to the classes that are present in the sequence. In this Figure \ref{fig:class_attention_individual}, we show attention maps for two instances from the same video and it can be observed that the active classes in the attention maps change with the actions present in the video.

\begin{figure*}[t!]
\begin{center}
\includegraphics[width=\linewidth]{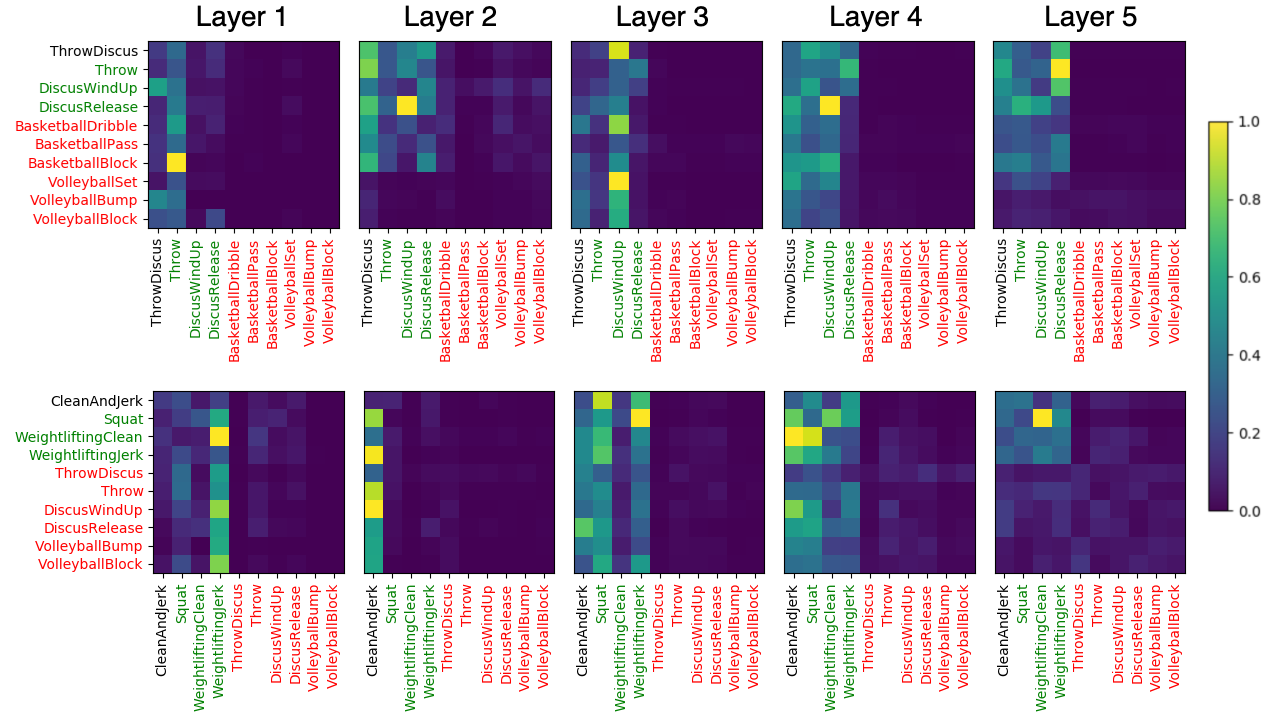}
\end{center}
\caption{We select 10 actions out of 65 from Multi-THUMOS, and visualize a $10\times10$ subsets of the CB attention maps from all MLAD layers obtained by averaging the maps over time-steps where action ``Throw Discus" (top) and ``Clean and Jerk" (bottom) are present. Columns corresponding to classes that are related to the main action (those in green) tend to be active, whereas unrelated classes (in red) tend to have low activation.} 
\label{fig:class_attention_complete}
\end{figure*}

\begin{figure*}[t!]
\centering
\includegraphics[width=\linewidth]{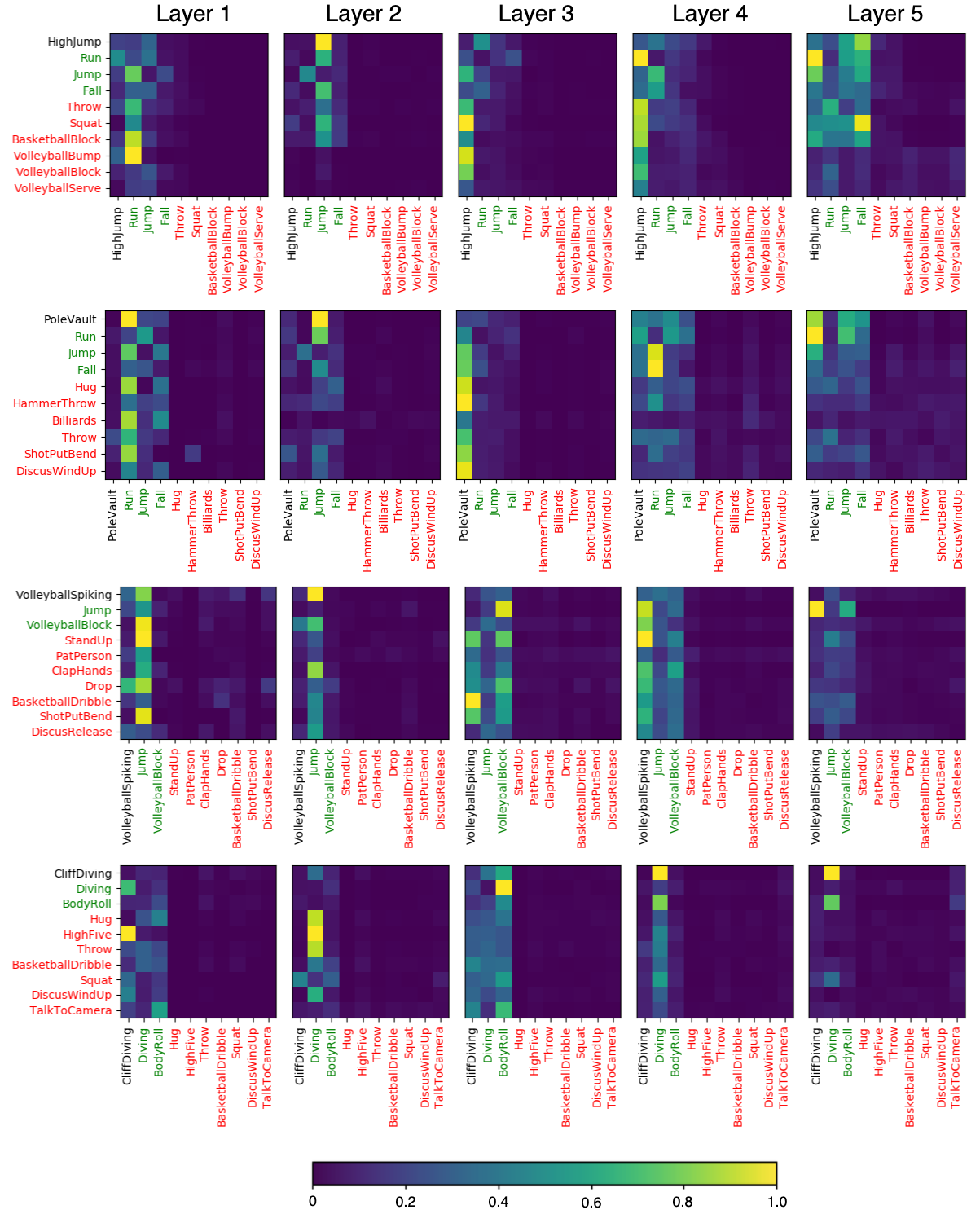}
\caption{Visualization of $10\times10$ subsets of the CB attention maps from all MLAD layers obtained by averaging the maps over time-steps where a specific action (shown in black) is present.  Columns corresponding to classes that are related to the main action (those in green) tend to be active, whereas unrelated classes (in red) tend to have low activation.} 
\label{fig:class_attention2_complete}
\end{figure*}

In Figure \ref{fig:class_attention_complete}, we present the class attention maps from all the five layers of our model. Attention maps from only layers 1, 3, and 5 are shown in the main paper. In Figure \ref{fig:class_attention2_complete}, we show class attention maps for more classes. Each row in the figure is for a specific class (main class, shown in black) and each column shows the attention map from a specific layer. In these attention maps, classes related to the main class (shown in green) tend to have more activation than the unrelated classes (shown in red).

\begin{figure*}[t!]
\begin{center}
\includegraphics[width=\linewidth]{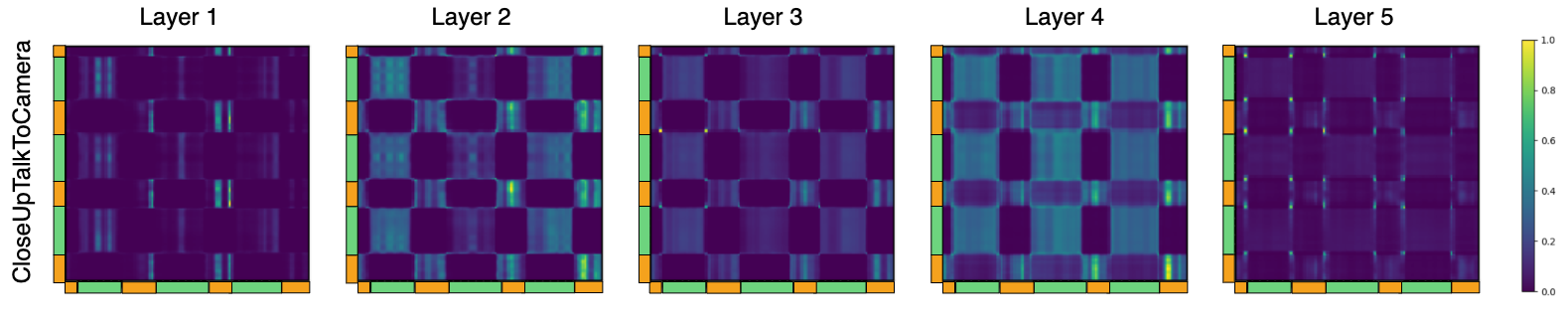}
\end{center}
\caption{Visualization of time attention maps from every MLAD layer of our network. Attention maps shown here are of dimension $T \times T$ ($T$ = 128). This shows the attention maps for class ``Close Up Talk To Camera" from a sample sequence.  Highlighted regions in green are the timesteps where the action appears in the sample and regions in orange are the timesteps without the action.
} 
\label{fig:time_attention_individual2}
\end{figure*}

\begin{figure*}[t!]
\begin{center}
\includegraphics[width=\linewidth]{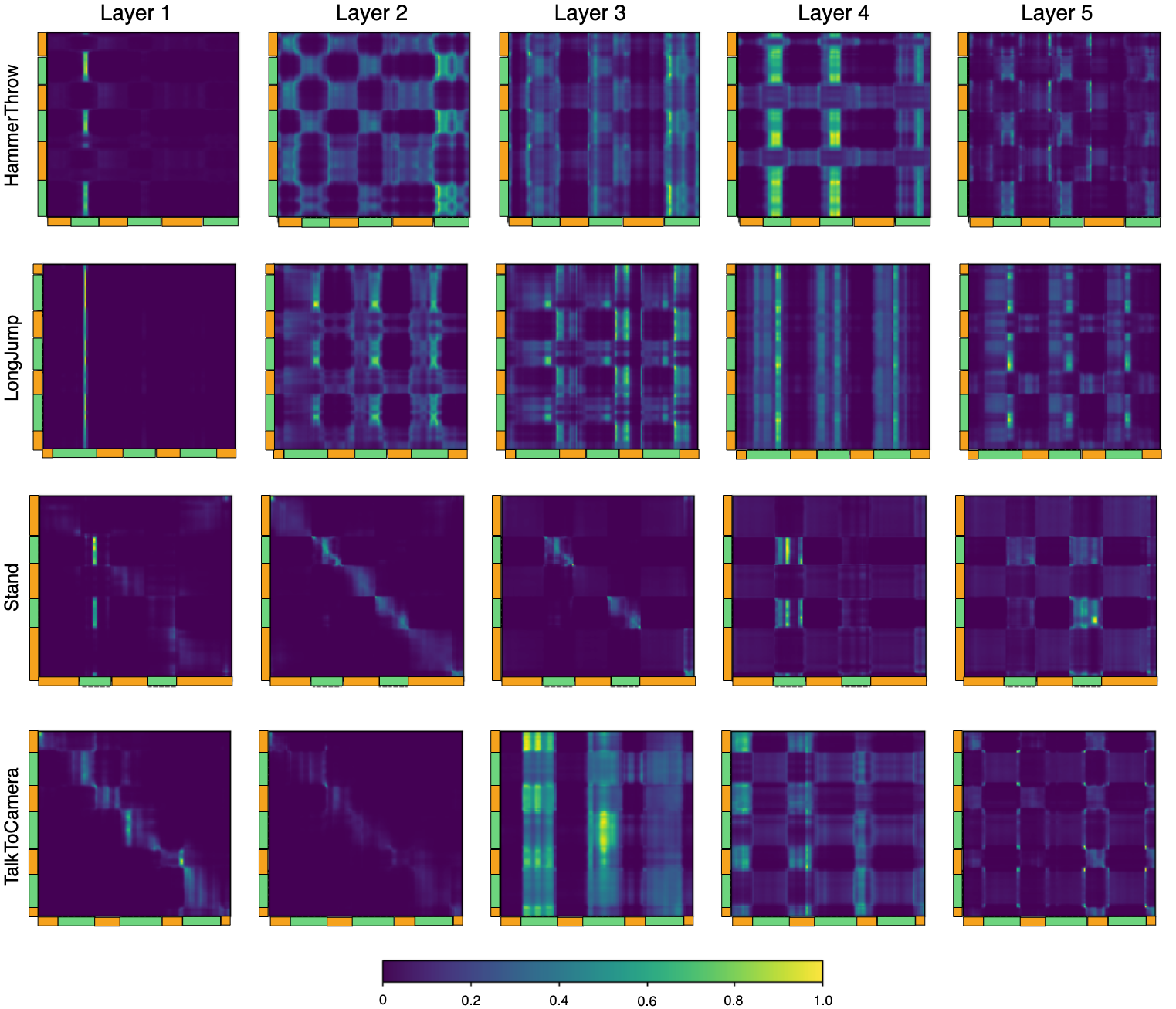}
\end{center}
\caption{Visualization of time attention maps from every MLAD layer of our network. Attention maps shown here are of dimension $T \times T$ ($T$ = 128). Highlighted regions in green are the timesteps where the action appears in the sample. Each row is for a specific class and the corresponding action class is shown on the left of each row.
} 
\label{fig:time_attention2_individual}
\end{figure*}

In Figure \ref{fig:time_attention_individual2}, we present the $T \times T$ attention maps from each MLAD layer for a specific class, ``Close Up Talk To Camera". Attention maps from only layers 1, 3, and 5 are presented in the main paper.  In Figure \ref{fig:time_attention2_individual}, we present the $T \times T$ attention maps from each MLAD layer for more classes in the MultiTHUMOS dataset. In all these samples, "Checkerboard" pattern is predominantly visible. This indicates that, timesteps where the action is present focus on other timesteps containing the action and vice-versa. We observe that role of each layer changes from one class to another. 

Other patterns are also common in these attention maps. For example, a layer may focus on time-steps near the current time-step with similar features, resulting in an attention matrix with activations along the main diagonal (see layer 2 for the ``Stand" action and layers 1 and 2 for the ``TalkToCamera" action). Also, a layer may focus on all time-steps in which the action occurs, leading to vertical bars (see layer 4 for the ``Long Jump" action and layer 3 for the ``TalkToCamera" action).

{\small
\bibliographystyle{ieee_fullname}
\bibliography{egbib}
}

\end{document}